\documentclass{article}
\usepackage{ijcai25}

\usepackage{times}
\usepackage{soul}
\usepackage{url}
\usepackage[hidelinks]{hyperref}
\usepackage[utf8]{inputenc}
\usepackage[small]{caption}
\usepackage{graphicx}
\usepackage{amsmath}
\usepackage{amsthm}
\usepackage{booktabs}
\usepackage{algorithm}
\usepackage{algorithmic}
\usepackage{xspace}
\usepackage{xcolor}
\usepackage[switch]{lineno}
\usepackage{multicol}
\usepackage{multirow}
\usepackage{array}
\usepackage{pdfrender}
\usepackage{colortbl}
\usepackage{makecell}
\usepackage{amssymb}
\usepackage{pifont}
\usepackage{ragged2e}
\newcommand{\boldcheckmark}{\textcolor{blue}{\ding{51}}}
\newcommand{\xmark}{\ding{55}}
\newcommand{\boldxmark}{\textcolor{purple}{{\textbf{\xmark}}}}

\usepackage[capitalize]{cleveref}
\crefname{section}{Sec.}{Secs.}
\Crefname{section}{Section}{Sections}
\Crefname{table}{Table}{Tables}
\crefname{table}{Table}{Tables}

\def\R{\mathbb{R}}
\newcommand{\C}[1]{\mathcal{#1}}
\newcommand{\bd}[1]{\boldsymbol{#1}}

\makeatletter
\DeclareRobustCommand\onedot{\futurelet\@let@token\@onedot}
\def\@onedot{\ifx\@let@token.\else.\null\fi\xspace}
\definecolor{LightCyan}{rgb}{0.88,1,1}

\makeatletter
\DeclareRobustCommand\onedot{\futurelet\@let@token\bmv@onedotaux}
\def\bmv@onedotaux{\ifx\@let@token.\else.\null\fi\xspace}
 
\def\ie{\emph{i.e}\onedot} 
 
\def\etc{\emph{etc}\onedot}

\makeatother

\title{A Survey of Pathology Foundation Model: Progress and Future Directions}

\author{
Conghao Xiong$^1$\and
Hao Chen$^{2}$\And
Joseph~J.~Y.~Sung$^3$
\affiliations
$^1$Department of Computer Science and Engineering, The Chinese University of Hong Kong\\
$^2$Department of Computer Science and Engineering and Department of Chemical and Biological Engineering, The Hong Kong University of Science and Technology\\
$^3$Lee Kong Chian School of Medicine, Nanyang Technological University
\emails
chxiong21@cse.cuhk.edu.hk,
jhc@cse.ust.hk,
josephsung@ntu.edu.sg
}

\begin{document}

\maketitle

\begin{abstract}
    Computational pathology, which involves analyzing whole slide images for automated cancer diagnosis, relies on multiple instance learning, where performance depends heavily on the feature extractor and aggregator. Recent Pathology Foundation Models (PFMs), pretrained on large-scale histopathology data, have significantly enhanced both the extractor and aggregator, but they lack a systematic analysis framework. In this survey, we present a hierarchical taxonomy organizing PFMs through a top-down philosophy applicable to foundation model analysis in any domain: model scope, model pretraining, and model design. Additionally, we systematically categorize PFM evaluation tasks into slide-level, patch-level, multimodal, and biological tasks, providing comprehensive benchmarking criteria. Our analysis identifies critical challenges in both PFM development (pathology-specific methodology, end-to-end pretraining, data-model scalability) and utilization (effective adaptation, model maintenance), paving the way for future directions in this promising field. Resources referenced in this survey are available at \url{https://github.com/BearCleverProud/AwesomeWSI}.
\end{abstract}
\section{Introduction}\label{sec:intro}
\textbf{C}omputational \textbf{Path}ology (CPath), the computational analysis of patient specimens (\ie, \textbf{W}hole \textbf{S}lide \textbf{I}mages, WSIs), is increasingly important due to the critical role of histopathology. For gigapixel WSIs, \textbf{M}ultiple \textbf{I}nstance \textbf{L}earning (MIL) is the de facto framework, involving WSI patch partitioning, feature extraction via pretrained neural networks, and feature aggregation into WSI-level features \cite{xiong2024takt}. Therefore, MIL performance hinges on two components: the pretrained neural network (\emph{extractor}) and the \emph{aggregator}.

\textbf{P}athology \textbf{F}oundation \textbf{M}odels (PFMs), neural networks pretrained on extensive pathological data that can be directly leveraged for diverse downstream tasks without retraining, such as HIPT~\cite{chen2022scaling} and UNI~\cite{chen2024uni}, mark a paradigm shift for MIL. Conventionally, due to the lack of PFMs, ResNet-50~\cite{he2016resnet} pretrained on ImageNet~\cite{deng2009imagenet} serves as the extractor~\cite{xiong2024mome}, but struggles with pathology-specific characteristics like minimal color variation, rotation-agnosticism, and hierarchical tissue organization. While limited labeled WSIs prevented supervised pretraining, \textbf{S}elf-\textbf{S}upervised \textbf{L}earning (SSL) enables PFMs that exhibit superior generalizability in morphology recognition. This overcomes natural image pretraining limitations, in which features mainly capture general visual attributes like edges and textures, enabling better performance on downstream tasks even with limited data.

Despite their potential, PFMs face multifaceted challenges: 1) most PFMs directly adopt natural image techniques, failing to cater to the discrepancy between pathology and natural images, indicating pathology-specific methodology remains underexplored; 2) MIL, as a two-stage pipeline, traps model training in local optima, while end-to-end training of WSIs requires prohibitive computational resources; 3) undefined model and data scaling bounds and resource constraints necessitate multi-institutional federated learning, demanding efficiency; and 4) the computational demands of PFMs impede deployment and maintenance, requiring continuous adaptation to evolving WSI technologies and pathological variants.

\begin{figure*}
    \centering
    \includegraphics[width=\linewidth]{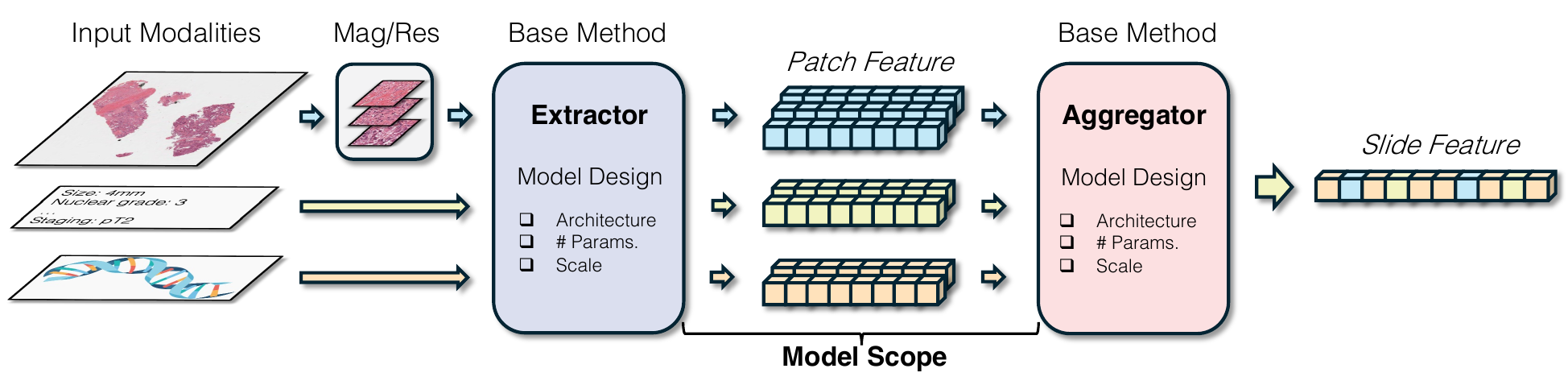}
    \caption{Schematic representation of our hierarchical taxonomy integrated within the MIL framework for PFMs.}
    \label{fig:mil}
\end{figure*}

Recent surveys on PFMs have contributed significantly to the understanding of the field; however, these works either focus primarily on the impact of PFMs on the real world rather than technical investigations of them~\cite{ochi2025pathology}, or detail the previous efforts in this field without a systematic taxonomy for technical analysis and a systematic organization of the evaluation tasks of PFMs~\cite{chanda2024new}. To address these critical gaps, we introduce a comprehensive and timely survey of the current landscape. We collected papers from high-impact journals, including Nature, Nature Medicine, Nature Biomedical Engineering, Medical Image Analysis, as well as top-tier conferences such as CVPR, ICML, and AAAI. Given the rapid evolution of the field, we also incorporated preprints from repositories such as arXiv, bioRxiv, and medRxiv, acknowledging that many influential works are still under review. In total, our survey includes 27 PFM papers, 12 of which are preprints that have not yet been accepted by peer-reviewed conferences or journals.

We present this survey with three primary contributions: 1) a hierarchical taxonomy organizing PFMs based on scope, training strategy, and design to enable holistic analysis, transferable to general vision FMs; 2) a comprehensive analysis of evaluation methodologies, examining their technical merits and limitations; and 3) a structured analysis of pathology-centric research challenges prioritizing underexplored directions. The manuscript is organized as follows: \Cref{sec:prelim} formally formulates MIL and SSL; \Cref{sec:taxonomy} introduces the proposed hierarchical taxonomy; \Cref{sec:benchmark} examines evaluation tasks for PFMs; \Cref{sec:challenges} delineates future research directions in this field; and \Cref{sec:conc} concludes our survey.

\section{Background and Problem Formulation}\label{sec:prelim}
\subsection{Multiple Instance Learning}\label{sec:mil}
In the MIL framework, a WSI is typically represented as a bag of $N$ unordered instances (or patches). The central objective of MIL is to predict the WSI-level label $\hat{Y}$ using only the ground truth bag label $Y$ as supervision, without access to the ground truth instance-level labels $\{y_i\}_{i=1}^{N}$. This setting reflects a common scenario in computational pathology, where obtaining slide-level annotations is feasible, but annotating individual patches is prohibitively expensive and impractical. The relationship between the bag label $Y$ and the instance labels $\{y_i\}_{i=1}^{N}$ is typically defined under standard MIL assumptions such as the presence-based assumption, and can be formally expressed as \cite{xiong2023diagnose},
\begin{equation}\label{eq:mil}
Y =
\begin{cases}
1 & \exists i, y_i=1\\
0 & \forall i, y_i=0\\
\end{cases}.
\end{equation}
The implementation of MIL involves tessellating WSIs into non-overlapping patches $\bd{X} = \{\bd{x}_i\}_{i=1}^{N} \in \R^{N\times h \times w \times 3}$, with $h,w$ standing for height and width, respectively. These patches undergo feature extraction through an extractor $\C{M}_e(\cdot)$, generating corresponding features $\bd{Z} = \{\bd{z}_i\}_{i=1}^{N} \in \R^{N\times d}$, where each feature is computed as $\bd{z}_i = \C{M}_e(\bd{x}_i)$, and $d$ is the hidden dimension of the extractor. Subsequently, an aggregator $\C{M}_g(\cdot)$ agglomerates these features to form a bag-level feature $\bd{h} = \C{M}_g({\bd{Z}})$ of the WSI, which finally serves as the input for the classification layer. Throughout aggregation, the extractor $\C{M}_e(\cdot)$ usually remains frozen and non-trainable due to GPU memory constraints, while the aggregation network $\C{M}_g(\cdot)$ is optimized during training. We refer readers unfamiliar with MIL to prior surveys for more details~\cite{carbonneau2018multiple,waqas2024exploring}.

\subsection{Self-supervised Learning}\label{sec:ssl}
SSL leverages unlabeled data by automatically generating supervisory signals through pretext tasks~\cite{ericsson2022self}. Given an input image $\bd{x}$, a transformation function $\mathcal{T}(\cdot)$ is applied to generate a modified version $\tilde{\bd{x}} = \mathcal{T}(\bd{x})$ and a corresponding pseudo-label $\tilde{y}$. 
An extractor $\C{M}_e(\cdot)$ extracts features from $\tilde{\bd{x}}$ and generates a predicted label $\hat{y} = \C{M}_e(\tilde{\bd{x}})$.
The learning objective can be formalized as minimizing the difference between the predicted label $\hat{y}$ and the pseudo-label $\tilde{y}$. Common pretext tasks include contrastive learning, self-distillation, masked image modeling, \etc, each designed to force the model to learn meaningful semantic features of the data. Through this process, the extractor can learn transferable features for downstream tasks on massive unlabeled data. We refer readers who are unfamiliar with SSL to prior surveys for more details~\cite{ericsson2022self,shurrab2022self,gui2024survey}.

\section{Hierarchical Taxonomy}\label{sec:taxonomy}
\newcolumntype{P}[1]{>{\centering\arraybackslash}p{#1}}
\begin{table*}[!ht]
    \small
    \centering
    \resizebox{\textwidth}{!}{%
    \begin{tabular}{c P{0.8cm}P{0.8cm} P{0.1cm} ccc P{0.1cm} ccc}
    \toprule
    \multirow{2}{*}{\textbf{Model}} &
  \multicolumn{2}{c}{\textbf{Model Scope}} & & 
  \multicolumn{3}{c}{\textbf{Model Pretraining}} & & 
   \multicolumn{3}{c}{\textbf{Model Design}} \\ 
   \cline{2-3} \cline{5-7} \cline{9-11}
 & \textbf{E.} & \textbf{A.} & & \textbf{Input} &\textbf{Base Method} &\textbf{Mag/Res} & &\textbf{Architecture} &\textbf{\# Params.} & \textbf{Scale}\\
    \midrule
    \rowcolor{gray!10}
    CTransPath & \boldcheckmark & \boldxmark & & H & MoCov3 & 10/224 & & Swin-T/14 & 28.3M & S\\
    REMEDIS & \boldcheckmark & \boldxmark & & H & SimCLR & Multi/224 & & ResNet-50 & 25.6M & S\\
    \rowcolor{gray!10}
    HIPT &\boldcheckmark & \boldcheckmark & & H & DINO & 20/256,4096 & & ViT-S/16-XS/256 & 21.7/2.78M & S/XS \\
    PLIP &\boldcheckmark & \boldxmark & & P, T & CLIP & 20/224 & & ViT-B/32 & 87M & B\\
    \rowcolor{gray!10}
    CONCH &\boldcheckmark & \boldxmark & & W, T & iBOT/CoCa & 20/256 & & ViT/B-16 & 86.3M & B\\
    Phikon &\boldcheckmark & \boldxmark & & H & iBOT & 20/224 & & ViT-S/B/L/16 & 21.7/85.8/307M & S/B/L\\
    \rowcolor{gray!10}
    UNI &\boldcheckmark & \boldxmark & & H & DINOv2 & 20/256,512 & & ViT-L/16 & 307M & L\\
    Virchow &\boldcheckmark & \boldxmark & & H & DINOv2 & 20/224 & & ViT-H/14 & 632M & H\\
    \rowcolor{gray!10}
    SINAI &\boldcheckmark & \boldxmark & & H & DINO/MAE & Unknown & & ViT-S/L & 21.7M/303.3M & S/L\\
    CHIEF & \boldxmark & \boldcheckmark & & H,T & Sup.+CLIP & 10/224 & & CHIEF & 1.2M & XS\\
    \rowcolor{gray!10}
    Prov-GigaPath &\boldcheckmark & \boldcheckmark & & H,I & DINOv2/MAE & 20/256 & & ViT-g/14/LongNet & 1.13B/85.1M & g/B\\
    Pathoduet &\boldcheckmark & \boldxmark & & H,I & MoCov3 & 40/256,20/1024 & & ViT-B/16 & 85.8M & B\\
    \rowcolor{gray!10}
    RudolfV & \boldcheckmark & \boldxmark & & W & DINOv2 & 20,40,80/256 & & ViT-L/14 & 304M & L\\
    PLUTO &\boldcheckmark & \boldxmark & & W & DINOv2 & 20,40/224 & & FlexiViT-S/16 & 22M & S\\
    \rowcolor{gray!10}
    PRISM &\boldxmark & \boldcheckmark & & H,T & CoCa & 20/224 & & Perceiver & 45.0M & S\\
    TANGLE &\boldcheckmark & \boldcheckmark & & H,G & iBOT/SimCLR & 20/224 & & ViT-B/16/ABMIL & 86.3/2.3M & B/XS\\
    \rowcolor{gray!10}
    MUSK & \boldcheckmark & \boldxmark &&  H,T & MIM & 10,20,40/384 & & BEiT-3 & 675M & H\\
    BEPH &\boldcheckmark & \boldxmark & & H & MIM & 40/224 & & BEiTv2 & 192.55M & B\\
    \rowcolor{gray!10}
    Hibou &\boldcheckmark & \boldxmark & & W & DINOv2 & Unknown & & ViT-B/L/16 & 86.3/307M & B/L\\
    mSTAR+ &\boldcheckmark & \boldcheckmark & & H,G,T & CLIP/ST & 20/256 & & TransMIL/ViT-L & 2.67/307M & XS/L\\
    \rowcolor{gray!10}
    GPFM &\boldcheckmark & \boldxmark & & H & UDK & 40/512 & & ViT-L/14 & 307M & L\\
    Virchow2G & \boldcheckmark & \boldxmark & & W & DINOv2 & 5,10,20,40/224 & & ViT-G/14 & 1.9B & G\\
    \rowcolor{gray!10}
    MADELEINE & \boldxmark & \boldcheckmark & & W & CLIP & 10,20/256 & & MH-ABMIL & 5.0M & XS\\
    Phikon-v2 & \boldcheckmark & \boldxmark & & W & DINOv2 & 20/224 & & ViT-L/16 & 307M & L\\
    \rowcolor{gray!10}
    TITAN &\boldxmark & \boldcheckmark & & W,T & iBOT/CoCa & 20/8192 & & TITAN/TITAN$_V$ & 48.5/42.1M & S\\
    KEEP &\boldcheckmark & \boldxmark & & W,T & CLIP & 20/224 & & UNI & 307M & L\\
    \rowcolor{gray!10}
    THREADS &\boldxmark & \boldcheckmark & & H,D,R & CLIP & 20/512 & & MH-ABMIL & 11.3M & XS\\
    \bottomrule
    \end{tabular}}
    \caption{Systematic comparison of PFMs categorized based on our hierarchical taxonomy. Abbreviations used: Extractor (E.), Aggregator (A.), H\&E (H), Patch (P), Text (T), WSIs with unspecified stains (W), IHC (I), Genomics (G), DNA (D), and RNA (R).}
    \label{tab:taxonomy}
\end{table*}
Our taxonomy systematically organizes PFMs through three interdependent dimensions and reflects a top-down design philosophy: 1) \emph{Model Scope}: a categorization of the scope of the PFMs, differentiating between PFMs focused on extractors, aggregators, and both components; 2) \emph{Model Pretraining}: a detailed examination of the spectrum of image-centric pretraining methods, including slide-level, patch-level, and multimodal techniques; and 3) \emph{Model Design}: a rigorous analysis of architecture, categorizing PFMs according to their number of parameters and scale. This top-down structure enables systematic comparisons of PFMs, as shown in \cref{tab:taxonomy}.

\subsection{Model Scope}\label{sec:scope}
MIL consists of three parts: 1) patch partitioning, 2) feature extraction, and 3) feature aggregation. As patch partitioning has been well-established, MIL performance primarily depends on the extractor and aggregator. In addition, WSIs inherently exhibit hierarchical structures, where local histomorphological patterns captured by extractors and global hierarchical tissue organization modeled by aggregators jointly determine diagnostic accuracy. Therefore, we categorize PFMs based on their scope: \textit{extractor-centric}, \textit{aggregator-centric}, or \textit{hybrid-centric}. The categorization of PFMs along this dimension is presented in the Model Scope column of \cref{tab:taxonomy}.

\textbf{Extractor-centric models} constitute the predominant approach in PFM development, driven by two factors: the importance of high-quality features and the necessity to address domain shift brought by ImageNet-pretrained CNNs. The role of the extractor aligns with established clinical practice, where pathologists emphasize cellular morphological analysis at the patch level. CTransPath~\cite{wang2022transformer} pioneers the extractor training with a hybrid CNN-Transformer design through \textbf{S}emantic-\textbf{R}elevant \textbf{C}ontrastive \textbf{L}earning (SRCL) on 15 million patches. REMEDIS~\cite{azizi2023robust} demonstrates that the feature extraction capability of ResNet-50 is constrained by domain shift across different medical imaging domains. Various advancements, including Virchow~\cite{vorontsov2024virchow} and SINAI~\cite{campanella2024sinai}, further stress the significance of robust extractors.

\textbf{Aggregator-centric models} play a vital role in slide-level tasks as they are the only trainable models under direct supervision of ground truth labels, yet they are relatively underexplored compared to the extractor. CHIEF~\cite{wang2024pathology}, leveraging supervised pretraining with the anatomical site to create an anatomy-aware aggregator, first demonstrates the efficacy of aggregator pretraining. More recent research like MADELEINE~\cite{jaume2025multistain}, TITAN~\cite{ding2024titan}, and THREAD~\cite{vaidya2025thread} utilizes multimodal data in aggregator pretraining with frozen patch features to enhance performance across downstream tasks. This paradigm shift reflects growing awareness that the aggregator critically impacts downstream task performance, particularly in low-resource clinical scenarios~\cite{xu2024gigapath}. This observation aligns with transfer learning principles, wherein pretraining on large-scale datasets effectively alleviates downstream data scarcity challenges. However, empirical evidence also reveals that the pretrained CHIEF aggregator occasionally performs worse than linear probing of the extractor~\cite{ding2024titan}, which is potentially attributable to the small model size when trained on a pretraining-scale dataset, or to the conflicts between domain bias and generic features. Consequently, further investigations are warranted to assess the advantages of pretrained larger aggregators.

\textbf{Hybrid-centric models} are PFMs that pretrain both the extractor and aggregator. Their advantage lies in full exploitation of the aggregator, as the aggregators can flexibly adapt to the extractor with pretraining-scale data. HIPT pioneers this approach through hierarchical pretraining of the first two layers of the extractor, excluding the last layer, which is substantiated through empirical performance. Similarly, Prov-GigaPath~\cite{xu2024gigapath} pretrains a ViT extractor and a LongNet~\cite{ding2023longnet} slide encoder; however, LongNet generates instance-level features rather than a single slide-level feature, necessitating integration of ABMIL~\cite{ilse2018attention} or non-parametric pooling strategies for slide-level tasks. TANGLE \cite{jaume2024transcriptomics} pretrains both a ViT feature extractor and a transcriptomics-guided ABMIL aggregator. Finally, mSTAR~\cite{xu2024multimodal} distinguishes itself as a fully-pretrained hybrid-centric model by an inverted pretraining sequence, contrasting with the conventional paradigm: first optimizing the multimodal aggregator, followed by pretraining the extractor with the aggregator.

Analysis of recent developments reveals two observations. First, research emphasis has progressively shifted from feature extractor pretraining toward aggregator pretraining, a transition potentially attributable to both the robust performance of existing extractors and the increasing awareness of aggregator significance, especially in limited-data scenarios. Second, current aggregators demonstrate a hierarchical dependency pattern, wherein each successive model builds upon the capabilities of prior models. For instance, TITAN utilizes features from CONCHv1.5, which in turn leverages UNI as its encoder, thereby forming a cascading performance dependency chain where the efficacy of TITAN is inherently contingent upon CONCHv1.5 and, by extension, UNI.

\subsection{Model Pretraining}
The pretraining methods can be categorized into supervised and SSL methods, with SSL prevailing due to their capabilities in capturing morphological patterns without labeled data, while only CHIEF opted for supervised pretraining for the aggregator. Based on our surveyed papers, SSL can be further divided into two main categories: vision-only and inter-modal methods. Vision-only methods employ three SSL techniques: \textit{contrastive learning} (SimCLR, MoCov3), \textit{masked image modeling} (MIM, MAE), and \textit{self-distillation} (iBOT, DINO, DINOv2). In contrast, inter-modal methods often employ multi-stage pretraining, utilizing contrastive learning methods (CLIP, CoCa) for effective cross-modal alignment before which unimodal encoders are pretrained independently. We focus on methodology contributions in this section and present the details of each method, including input modalities, magnification, and resolution of the patches, in the Model Pretraining column of \cref{tab:taxonomy}.

\textbf{Contrastive Learning} is an SSL branch that learns representations by maximizing similarity between positive pairs while minimizing that between negative pairs. Several seminal approaches have advanced this field: 1) SimCLR~\cite{chen2020simple} established foundational techniques such as aggressive data augmentation and large batch sizes; 2) MoCov3~\cite{chen2021empirical} advanced self-supervised learning for ViT through stabilized training techniques; 3) CLIP~\cite{radford2021clip} expanded the paradigm to multi-modal learning through large-scale image-caption pair training; and 4) CoCa~\cite{yu2022coca} proposed a unified method incorporating both contrastive and captioning objectives, enabling simultaneous visual-textual alignment and text generation capabilities. In the medical domain, REMEDIS utilizes SimCLR to enhance the robustness and data efficiency in medical imaging. TANGLE adopts a revised SimCLR method with gene expression reconstruction and slide subset alignment. Pathoduet~\cite{hua2024pathoduet} enhanced MoCov3 through the integration of cross-scale positioning and cross-stain transferring tasks, specifically addressing the challenges of stain transferability and tissue-level heterogeneity. CLIP is adapted for both extractors (PLIP~\cite{huang2023plip}) and aggregators (Prov-GiGapath, mSTAR, MADELEINE, and THREAD), due to its versatility in aligning two or more modalities. Notably, KEEP~\cite{zhou2024knowledge} has proposed a \textbf{K}nowledge-\textbf{E}nhanced \textbf{V}ision-\textbf{L}anguage (KEVL) pretraining, further adapting CLIP for the extractor by incorporating domain expertise through knowledge-graph-cleaned image-text pairs.
There are several applications of CoCa, both on the extractor and aggregator: CONCH~\cite{lu2024conch} adopts this framework to pretrain an extractor on 1.17 million image-caption pairs, enhancing both zero- and few-shot capabilities, while PRISM~\cite{shaikovski2024prism} and TITAN utilize CoCa to pretrain aggregators with multimodal capabilities. 

\textbf{Masked Image Modeling} is an SSL method that learns representations by predicting masked portions of images from their visible regions. SimMIM~\cite{xie2022simmim} advanced the field by simplifying existing approaches through random masking and a lightweight prediction head, and MAE~\cite{he2022masked} introduced an asymmetric encoder-decoder design with high masking ratios. Recent investigations have demonstrated the efficacy of MIM in pretraining extractors; notably, SINAI~\cite{campanella2024sinai} employs MAE to pretrain ViT models on a scale of 3.2 billion patches, establishing its scalability in pathological contexts. Similarly, MUSK~\cite{xiang2025musk} and BEPH~\cite{yang2024beph} further validate MIM by implementing BEiT-3 and BEiTv2 architectures, respectively. Additionally, Prov-GigaPath employs MAE to pretrain its slide encoder LongNet, demonstrating the efficacy of this method on aggregator pretraining. 

\textbf{Self-distillation} enables model learning through its own predictions across different views, simultaneously acting as teacher and student. DINO~\cite{caron2021emerging} pioneered the use of self-distillation by employing a teacher-student architecture with momentum encoder and multi-crop training, while iBOT~\cite{zhou2022image} performs MIM via self-distillation with an online tokenizer, and DINOv2~\cite{oquab2023dinov2} refined the DINO framework by accelerating and stabilizing the training at scale. The efficacy of self-distillation for the extractor has been demonstrated by several investigations: Phikon~\cite{filiot2023scaling} implements iBOT on a corpus of 43 million patches spanning 16 distinct cancer sites; Phikon-v2 \cite{filiot2024phikon} employs DINOv2 on 456 million patches derived from 30 cancer sites; RudolfV~\cite{dippel2024rudolfv} incorporates DINOv2 with pathologist knowledge on 58 tissue types and 129 stains; and Hibou~\cite{nechaev2024hibou} further extends DINOv2 on 1.2 billion patches. Additionally, the application of self-distillation extends beyond the extractor, as evidenced by TITAN~\cite{ding2024titan}, which utilizes iBOT for general-purpose aggregator learning. These investigations demonstrate the capacity of self-distillation in PFM pretraining. In addition, there are methodological improvements customized for pathology in this category: 1) PLUTO~\cite{juyal2024pluto} utilizes DINOv2 together with MAE objective and Fourier losses on 195 million patches; 2) GPFM~\cite{ma2024towards} proposes \textbf{U}nified \textbf{K}nowledge \textbf{D}istillation (UKD), incorporating MIM, self-distillation and expert knowledge distillation together as training objectives; 3) Virchow2~\cite{zimmermann2024virchow2} enhances DINOv2 by applying pathology-specific augmentation and reducing tissue redundancy.

\newcolumntype{L}[1]{>{\RaggedRight\arraybackslash}p{#1}}
\begin{table*}[!t]
    \centering
    \resizebox{\textwidth}{!}{%
    \begin{tabular}{L{5.57cm}P{2.08cm}P{2.25cm}P{2.27cm}P{2.07cm}P{3.78cm}c}
    \toprule
    \centering \bf Venue & \bf Model & \bf Method & \bf Architecture & \bf Data Source & \bf Data Statistics & \bf Links \\
    \midrule
    \rowcolor{gray!10}
    \textcolor{purple}{MedIA \cite{wang2022transformer}} & CTransPath & SRCL & Swin-T/14 & TCGA + PAIP & \makecell{32,220 WSIs \\ 15,580,262 Patches} & \href{https://github.com/Xiyue-Wang/TransPath}{\includegraphics[height=1em]{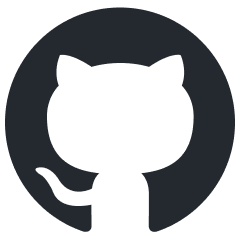}} \href{https://www.sciencedirect.com/science/article/pii/S1361841522002043}{\includegraphics[height=1em]{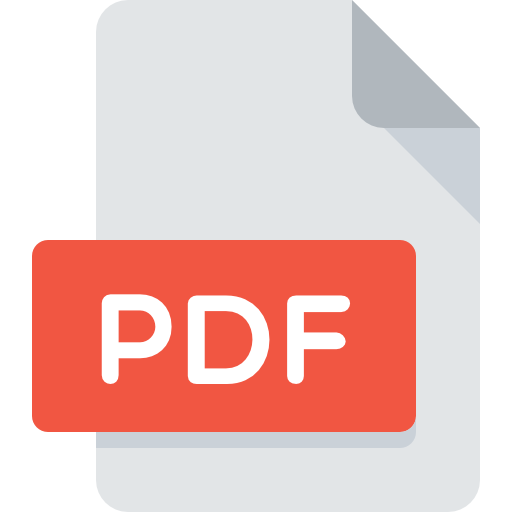}}\\
    \textcolor{purple}{Nat. Bio. Engg. \cite{azizi2023robust}} & REMEDIS & SimCLR & ResNet-50 & TCGA & \makecell{29,018 WSIs \\ 50 Million Patches} & \href{https://www.nature.com/articles/s41551-023-01049-7}{\includegraphics[height=1em]{figs/pdf.png}}\\
    \rowcolor{gray!10}
    \textcolor{purple}{CVPR \cite{chen2022scaling}} & HIPT & DINO & \makecell{ViT-S/16\\ViT-XS/256} & TCGA & \makecell{10,678 H\&E WSIs \\ $\sim$ 104 Million Patches} & \href{https://github.com/mahmoodlab/HIPT}{\includegraphics[height=1em]{figs/github-mark.png}} \href{https://openaccess.thecvf.com/content/CVPR2022/papers/Chen_Scaling_Vision_Transformers_to_Gigapixel_Images_via_Hierarchical_Self-Supervised_Learning_CVPR_2022_paper.pdf}{\includegraphics[height=1em]{figs/pdf.png}}\\
    \textcolor{purple}{Nat. Med. \cite{huang2023plip}} & PLIP & CLIP & ViT-B/32 & \makecell{OpenPath} & \makecell{208,414 Image-Text Pairs} & \href{https://huggingface.co/vinid/plip}{\includegraphics[height=1em]{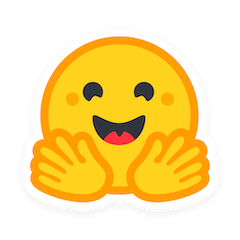}} \href{https://github.com/PathologyFoundation/plip}{\includegraphics[height=1em]{figs/github-mark.png}} \href{https://www.nature.com/articles/s41591-023-02504-3}{\includegraphics[height=1em]{figs/pdf.png}}\\
    \rowcolor{gray!10}
    \textcolor{purple}{Nat. Med. \cite{lu2024conch}} & CONCH & \makecell{P: iBOT\\A: CoCa} & \makecell{P: ViT-B/16\\A: GPT-style} & In-house & \makecell{21,442 WSIs \\ 16 Million Patches \\$>$ 1.17M Image-Text Pairs} & \href{https://huggingface.co/MahmoodLab/CONCH}{\includegraphics[height=1em]{figs/hf-logo.png}} \href{https://github.com/mahmoodlab/CONCH}{\includegraphics[height=1em]{figs/github-mark.png}} \href{https://www.nature.com/articles/s41591-024-02856-4}{\includegraphics[height=1em]{figs/pdf.png}}\\
    MedRxiv \cite{filiot2023scaling} & Phikon & iBOT & \makecell{ViT-S/B/L/16} & TCGA & \makecell{6,093 WSIs \\ 43,374,634 Patches} & \href{https://huggingface.co/owkin/phikon}{\includegraphics[height=1em]{figs/hf-logo.png}} \href{https://github.com/owkin/HistoSSLscaling?tab=readme-ov-file}{\includegraphics[height=1em]{figs/github-mark.png}} \href{https://www.medrxiv.org/content/10.1101/2023.07.21.23292757v2.full.pdf}{\includegraphics[height=1em]{figs/pdf.png}}\\
    \rowcolor{gray!10}
    \textcolor{purple}{Nat. Med. \cite{chen2024uni}} & UNI & DINOv2 & ViT-L/16 & Mass-100K & \makecell{100,426 H\&E WSIs \\ 100,130,900 Patches} &  \href{https://huggingface.co/MahmoodLab/UNI}{\includegraphics[height=1em]{figs/hf-logo.png}} \href{https://github.com/mahmoodlab/UNI/}{\includegraphics[height=1em]{figs/github-mark.png}} \href{https://www.nature.com/articles/s41591-024-02857-3}{\includegraphics[height=1em]{figs/pdf.png}} \\
    \textcolor{purple}{Nat. Med. \cite{vorontsov2024virchow}} & Virchow & DINOv2 & ViT-H/14 & MSKCC & \makecell{1,488,550 H\&E WSIs\\2 Billion Patches} & \href{https://huggingface.co/paige-ai/Virchow}{\includegraphics[height=1em]{figs/hf-logo.png}} \href{https://github.com/Paige-AI/paige-ml-sdk}{\includegraphics[height=1em]{figs/github-mark.png}} \href{https://www.nature.com/articles/s41591-024-03141-0}{\includegraphics[height=1em]{figs/pdf.png}}\\
    \rowcolor{gray!10}
    \textcolor{purple}{AAAI S. \cite{campanella2024sinai}} & SINAI & \makecell{DINO\\MAE} & \makecell{ViT-S\\ViT-L} & \makecell{Mount Sinai\\Health System} & \makecell{423,563 H\&E WSIs\\3.2 Billion Patches} & \href{https://github.com/fuchs-lab-public/OPAL/tree/main/SinaiPathologyFoundationModels}{\includegraphics[height=1em]{figs/github-mark.png}} \href{https://arxiv.org/pdf/2310.07033}{\includegraphics[height=1em]{figs/pdf.png}}\\
    \textcolor{purple}{Nature \cite{wang2024pathology}} & CHIEF & \makecell{P: Pretrained\\S: Sup.+CLIP} & \makecell{P: CTransPath\\S: CHIEF} & \makecell{Public +\\ In-house} & \makecell{60,530 H\&E WSIs \\ $\sim$ 15 Million Patches} &  \href{https://hub.docker.com/r/chiefcontainer/chief/}{\includegraphics[height=1em]{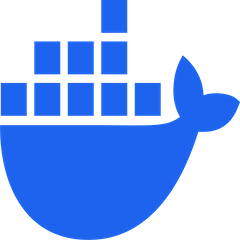}} \href{https://github.com/hms-dbmi/chief}{\includegraphics[height=1em]{figs/github-mark.png}} \href{https://www.nature.com/articles/s41586-024-07894-z}{\includegraphics[height=1em]{figs/pdf.png}}\\
    \rowcolor{gray!10}
    \textcolor{purple}{Nature \cite{xu2024gigapath}} & Prov-GigaPath & \makecell{P: DINOv2\\S: MAE\\A: CLIP} & \makecell{P: ViT-g/14\\S: LongNet\\} & \makecell{Providence \\ Health System} & \makecell{171,189 WSIs \\ 1,384,860,229 Patches} &  \href{https://huggingface.co/prov-gigapath/prov-gigapath}{\includegraphics[height=1em]{figs/hf-logo.png}} \href{https://github.com/prov-gigapath/prov-gigapath}{\includegraphics[height=1em]{figs/github-mark.png}} \href{https://www.nature.com/articles/s41586-024-07441-w}{\includegraphics[height=1em]{figs/pdf.png}}\\  
    \textcolor{purple}{MedIA \cite{hua2024pathoduet}} & Pathoduet & \makecell{Enhanced\\MoCov3} & \makecell{ViT-B/16} & TCGA & \makecell{11,000 WSIs \\ 13,166,437 Patches} & \href{https://github.com/openmedlab/PathoDuet}{\includegraphics[height=1em]{figs/github-mark.png}} \href{https://www.sciencedirect.com/science/article/pii/S1361841524002147}{\includegraphics[height=1em]{figs/pdf.png}}\\
    \rowcolor{gray!10}
    Arxiv \cite{dippel2024rudolfv} & RudolfV & DINOv2 & ViT-L/14 & \makecell{TCGA +\\In-house}& \makecell{133,998 WSIs\\1.25 Billion Patches} & \href{https://arxiv.org/pdf/2401.04079}{\includegraphics[height=1em]{figs/pdf.png}} \\
    \textcolor{purple}{ICML W. \cite{juyal2024pluto}} & PLUTO & \makecell{DINOv2+\\MAE+Fourior} & \makecell{FlexiViT-S/16} & \makecell{TCGA +\\Proprietary} & \makecell{158,852 WSIs \\ 195 Million Patches} & \href{https://arxiv.org/pdf/2405.07905}{\includegraphics[height=1em]{figs/pdf.png}} \\
    \rowcolor{gray!10}
    Arxiv \cite{shaikovski2024prism} & PRISM & \makecell{P: Pretrained\\S: CoCa} & \makecell{P: Virchow\\S: Perceiver} & MSKCC & \makecell{587,196 WSIs\\195K Pathology Reports} & \href{https://huggingface.co/paige-ai/Prism}{\includegraphics[height=1em]{figs/hf-logo.png}} \href{https://arxiv.org/pdf/2405.10254}{\includegraphics[height=1em]{figs/pdf.png}} \\
    \textcolor{purple}{CVPR \cite{jaume2024transcriptomics}} & TANGLE & \makecell{P: iBOT\\S: Alignment} & \makecell{P: ViT-B/16\\S: ABMIL} & \makecell{TG-GATEs} & \makecell{47,227 WSIs\\6,597 Image-Gene Pair} & \href{https://github.com/mahmoodlab/TANGLE}{\includegraphics[height=1em]{figs/github-mark.png}} \href{https://openaccess.thecvf.com/content/CVPR2024/papers/Jaume_Transcriptomics-guided_Slide_Representation_Learning_in_Computational_Pathology_CVPR_2024_paper.pdf}{\includegraphics[height=1em]{figs/pdf.png}} \\
    \rowcolor{gray!10}
    \textcolor{purple}{Nature \cite{xiang2025musk}} & MUSK & UMP & BEIT-3 & \makecell{Quilt-1M +\\ PathAsst} & \makecell{$\sim$33,000 H\&E WSIs\\50M Patches\\1M Image-Text Pairs}& \href{https://huggingface.co/xiangjx/musk}{\includegraphics[height=1em]{figs/hf-logo.png}} \href{https://github.com/lilab-stanford/MUSK}{\includegraphics[height=1em]{figs/github-mark.png}} \href{https://www.nature.com/articles/s41586-024-08378-w}{\includegraphics[height=1em]{figs/pdf.png}}\\
    BioRxiv \cite{yang2024beph} & BEPH & MIM & BEiTv2 & TCGA & \makecell{11,760 WSIs\\11,774,353 Patches}& \href{https://github.com/Zhcyoung/BEPH}{\includegraphics[height=1em]{figs/github-mark.png}} \href{https://www.biorxiv.org/content/10.1101/2024.05.16.594499v1.full}{\includegraphics[height=1em]{figs/pdf.png}}\\
    \rowcolor{gray!10}
    Arxiv \cite{nechaev2024hibou} & Hibou & DINOv2 & \makecell{ViT-L/14\\ViT-B/14}& Proprietary & \makecell{936,441 H\&E WSIs\\202,464 non-H\&E WSIs\\ViT-L: 1.2B Patches\\ViT-B: 512M Patches}& \href{https://huggingface.co/histai/hibou-L}{\includegraphics[height=1em]{figs/hf-logo.png}} \href{https://github.com/HistAI/hibou}{\includegraphics[height=1em]{figs/github-mark.png}} \href{https://arxiv.org/pdf/2406.05074}{\includegraphics[height=1em]{figs/pdf.png}} \\
    Arxiv \cite{xu2024multimodal} & mSTAR+ & \makecell{S: CLIP\\P: mSTAR} & \makecell{S: TransMIL\\P: ViT-L} & TCGA & \makecell{11,727 WSIs\\22,127 Modality Pairs} & \href{https://huggingface.co/Wangyh/mSTAR}{\includegraphics[height=1em]{figs/hf-logo.png}} \href{https://github.com/Innse/mSTAR}{\includegraphics[height=1em]{figs/github-mark.png}} \href{https://arxiv.org/pdf/2407.15362}{\includegraphics[height=1em]{figs/pdf.png}} \\
    \rowcolor{gray!10}
    Arxiv \cite{ma2024towards} & GPFM & \makecell{UKD} & ViT-L/14& \makecell{33 Public\\Dataset} & \makecell{72,280 WSIs\\190,212,668 Patches}& \href{https://huggingface.co/majiabo/GPFM}{\includegraphics[height=1em]{figs/hf-logo.png}} \href{https://github.com/birkhoffkiki/GPFM}{\includegraphics[height=1em]{figs/github-mark.png}} \href{https://arxiv.org/pdf/2407.18449}{\includegraphics[height=1em]{figs/pdf.png}} \\
    Arxiv \cite{zimmermann2024virchow2} & \makecell{Virchow2\\Virchow2G} & \makecell{Enhanced\\DINOv2}& \makecell{ViT-H/14\\ViT-G/14}& \makecell{MSKCC +\\Worldwide} & \makecell{3,134,922 WSIs\\with Diverse Stains} & \href{https://huggingface.co/paige-ai/Virchow2}{\includegraphics[height=1em]{figs/hf-logo.png}} \href{https://arxiv.org/pdf/2408.00738}{\includegraphics[height=1em]{figs/pdf.png}} \\
    \rowcolor{gray!10}
    \textcolor{purple}{ECCV \cite{jaume2025multistain}} & MADELEINE & \makecell{P: Pretrained\\S: CLIP + GOT}& \makecell{P: CONCH\\S:MH-ABMIL}& \makecell{Acrobat +\\BWH} & \makecell{16,281 WSIs\\with Diverse Stains}& \href{https://huggingface.co/MahmoodLab/madeleine}{\includegraphics[height=1em]{figs/hf-logo.png}} \href{https://github.com/mahmoodlab/MADELEINE}{\includegraphics[height=1em]{figs/github-mark.png}} \href{https://link.springer.com/chapter/10.1007/978-3-031-73414-4_2}{\includegraphics[height=1em]{figs/pdf.png}} \\
    Arxiv \cite{filiot2024phikon} & Phikon-v2 & DINOv2 & ViT-L/16 & \makecell{Public + \\In-house} & \makecell{58,359 WSIs\\ 456,060,584 Patches} & \href{https://huggingface.co/owkin/phikon-v2}{\includegraphics[height=1em]{figs/hf-logo.png}} \href{https://arxiv.org/pdf/2409.09173}{\includegraphics[height=1em]{figs/pdf.png}}\\
    \rowcolor{gray!10}
    Arxiv \cite{ding2024titan} & TITAN & \makecell{P: Pretrained\\Stage1: iBOT\\Stage2: CoCa} & \makecell{P: CONCHv1.5\\S: ViT-T/14} & Mass-340K & \makecell{335,645 WSIs \\ 423,122 Image-Text Pairs \\ 182,862 WSI-Text Pairs} &  \href{https://huggingface.co/MahmoodLab/TITAN}{\includegraphics[height=1em]{figs/hf-logo.png}} \href{https://github.com/mahmoodlab/TITAN}{\includegraphics[height=1em]{figs/github-mark.png}} \href{https://arxiv.org/pdf/2411.19666}{\includegraphics[height=1em]{figs/pdf.png}} \\
    Arxiv \cite{zhou2024knowledge} & KEEP & KEVL & UNI & \makecell{Quilt-1M + \\OpenPath}& \makecell{143K Image-Text Pairs\\Hierarchical Medical KG}& \href{https://huggingface.co/Astaxanthin/KEEP}{\includegraphics[height=1em]{figs/hf-logo.png}} \href{https://github.com/MAGIC-AI4Med/KEEP}{\includegraphics[height=1em]{figs/github-mark.png}} \href{https://arxiv.org/pdf/2412.13126}{\includegraphics[height=1em]{figs/pdf.png}}\\
    \rowcolor{gray!10}
    Arxiv \cite{vaidya2025thread} & THREADS & \makecell{P: Pretrained\\S: CLIP\\} & \makecell{P: CONCHv1.5\\S: MH-ABMIL}& \makecell{MBTG-47K:\\MGH+BWH\\+TCGA\\+GTEx}& \makecell{47,171 H\&E WSIs\\125,148,770 Patches\\26,615 Bulk RNA\\20,556 DNA Variants}& \href{https://huggingface.co/datasets/MahmoodLab/Patho-Bench}{\includegraphics[height=1em]{figs/hf-logo.png}} \href{https://github.com/mahmoodlab/patho-bench}{\includegraphics[height=1em]{figs/github-mark.png}} \href{https://arxiv.org/pdf/2501.16652}{\includegraphics[height=1em]{figs/pdf.png}}\\
    \bottomrule
    \end{tabular}
    }
    \caption{Technical specifications of vision-related parts of PFMs by academic preprint release date, with peer-reviewed published works in purple. Abbreviations used: Patch-level extractor (P), Alignment (A), Slide-level aggregator (S).}
    \label{tab:fms}
\end{table*}
\subsection{Model Design}
The model design refers to the following three aspects that are vital to model performance: \textit{architecture}, \textit{number of parameters} (\textit{\# params.}), \textit{scale}. The scale of a model is directly determined by its number of parameters. Through quantization of the number of parameters, we establish a hierarchical scale system, facilitating standardized cross-architectural comparisons and enabling informed model selection for practical implementations. The taxonomy of PFMs along this dimension is presented in the Model Design column of \cref{tab:taxonomy}.

\textbf{Architecture} is a pivotal determinant of PFM capabilities. Architectures adopted by PFMs can be categorized based on scope: extractor architecture and aggregator architecture. The extractor architecture encompasses two categories: 1) CNN-based architecture such as ResNet, and 2) transformer-based architecture, including ViT~\cite{dosovitskiy2021vit}, CNN-integrated Swin Transformer~\cite{wang2022transformer}, BEiTv2~\cite{peng2022beit}, FlexiViT~\cite{beyer2023flexivit}, and a multimodal BEiT-3~\cite{wang2023beit3}. The aggregator architecture comprises two categories: ABMIL family~\cite{ilse2018attention,ding2024titan} and Perceiver~\cite{jaegle2021perceiver}. For an architectural overview of each method, we refer readers to the Architecture column in \cref{tab:taxonomy}.

\textbf{Scale} can be derived through the number of parameters. To facilitate cross-architectural comparisons, we establish a quantization framework based on the ViT architecture, which serves as the predominant backbone across our surveyed literature. The classification includes seven categories: e\textbf{x}tra \textbf{s}mall (XS, 2.78M), \textbf{S}mall (S, 21.7M), \textbf{B}ase (B, 86.3M), \textbf{L}arge (L, 307M), \textbf{H}uge (H, 632M), \textbf{g}iant (g, 1.13B), and \textbf{G}iant (G, 1.9B). The notation ViT-B/16 indicates a ViT Base model with patch size 16. For statistics of each method, we refer readers to the \# Params. and Scale columns in \cref{tab:taxonomy}.

Our analysis reveals several patterns in model architectures and scaling: 1) ABMIL-derived methods demonstrate clear dominance in aggregator architectures, while the ViT family predominates in extractor architectures, which are transformer-based architectures; 2) The majority of methods utilize ViT-L as their primary backbone. Due to computational resource constraints, researchers often develop complementary smaller-scale variants (ViT-S or ViT-B) alongside their primary models, while some approaches specifically target efficiency through smaller architectures; 3) ViT-L is a popular scale for extractors, whereas ViT-XS is the primary choice for aggregators, over ViT-S. This substantial disparity in parameter counts between extractors and aggregators, despite their similar training data scale, suggests a potential data-model scale mismatch that warrants further investigation; 4) a clear trend toward larger model scales is observed: while earlier approaches frequently employed ViT-B, recent methods have increasingly standardized on ViT-L, with some extending to even larger variants such as ViT-H/g/G.

\begin{table*}[!ht]
    \small
    \centering
    \begin{tabular}{c cccc cccc cccc cccc}
    \toprule
    \multirow{2}{*}{\textbf{Model}} &
   \multicolumn{4}{c}{\textbf{Slide Level}} & & 
   \multicolumn{3}{c}{\textbf{Patch Level}} & & 
   \multicolumn{4}{c}{\textbf{Multimodal}} & & 
   \multicolumn{2}{c}{\textbf{Biological}}\\ 
   \cline{2-5} \cline{7-9} \cline{11-14} \cline{16-17}
 & \textbf{Cls.} & \textbf{Surv.} & \textbf{Retri.}& \textbf{Seg.}& & \textbf{Cls.} &\textbf{P2P} &\textbf{Seg.} & &\textbf{I2T} &\textbf{T2I} & \textbf{RG} & \textbf{VQA} & & \textbf{GA} & \textbf{MP}\\
    \midrule
    \rowcolor{gray!10}
    CTransPath & C & C & \boldxmark & \boldxmark & & F/C & Z & C & & \boldxmark & \boldxmark & \boldxmark & \boldxmark & & \boldxmark & \boldxmark \\
    REMEDIS & C & C & \boldxmark & \boldxmark & & \boldxmark & \boldxmark & \boldxmark & & \boldxmark & \boldxmark & \boldxmark & \boldxmark & & \boldxmark & \boldxmark\\
    \rowcolor{gray!10}
    HIPT & C & C & \boldxmark & \boldxmark & & \boldxmark & \boldxmark & \boldxmark & & \boldxmark & \boldxmark & \boldxmark & \boldxmark & & \boldxmark & \boldxmark\\
    PLIP & \boldxmark & \boldxmark & \boldxmark & \boldxmark & & Z & Z & \boldxmark & & \boldxmark & Z & \boldxmark & \boldxmark & & \boldxmark & \boldxmark\\
    \rowcolor{gray!10}
    CONCH & Z/F/C & \boldxmark & \boldxmark & Z & & Z/F & \boldxmark & \boldxmark & & Z & Z & C & \boldxmark & & \boldxmark & \boldxmark \\
    Phikon & C & C & \boldxmark & \boldxmark & & C & \boldxmark & \boldxmark & & \boldxmark & \boldxmark & \boldxmark & \boldxmark & & C & C\\
    \rowcolor{gray!10}
    UNI & F/C & \boldxmark & F & \boldxmark & & F/C & Z & C & & \boldxmark & \boldxmark & \boldxmark & \boldxmark & & \boldxmark & \boldxmark\\
    Virchow & C & \boldxmark & \boldxmark & \boldxmark & & C & \boldxmark & \boldxmark & & \boldxmark & \boldxmark & \boldxmark & \boldxmark & & C & \boldxmark\\
    \rowcolor{gray!10}
    SINAI & C & \boldxmark & \boldxmark & \boldxmark & & \boldxmark & \boldxmark & \boldxmark & & \boldxmark & \boldxmark & \boldxmark & \boldxmark & & C & C\\
    CHIEF & C & C & \boldxmark & \boldxmark & & \boldxmark & \boldxmark & \boldxmark & & \boldxmark & \boldxmark & \boldxmark & \boldxmark & & C & C\\
    \rowcolor{gray!10}
    Prov-GigaPath & Z/C & \boldxmark & C & \boldxmark & & \boldxmark & \boldxmark & \boldxmark & & \boldxmark & \boldxmark & \boldxmark & \boldxmark & & Z/C & \boldxmark \\
    Pathoduet & C & \boldxmark & \boldxmark & \boldxmark & & F/C & \boldxmark & \boldxmark & & \boldxmark & \boldxmark & \boldxmark & \boldxmark & & \boldxmark & F/C\\
    \rowcolor{gray!10}
    RudolfV & \boldxmark & \boldxmark & Z & \boldxmark & & C & \boldxmark & C & & \boldxmark & \boldxmark & \boldxmark & \boldxmark & & C & C\\
    PLUTO & C & \boldxmark & \boldxmark & \boldxmark & & C & \boldxmark & C & & \boldxmark & \boldxmark & \boldxmark & \boldxmark & & \boldxmark & C\\
    \rowcolor{gray!10}
    PRISM & Z/C & \boldxmark & \boldxmark & \boldxmark & & \boldxmark & \boldxmark & \boldxmark & & \boldxmark & \boldxmark & C & \boldxmark & & F/C & \boldxmark \\
    TANGLE & F & \boldxmark & C & \boldxmark & & \boldxmark & \boldxmark & \boldxmark & & \boldxmark & \boldxmark & \boldxmark & \boldxmark & & \boldxmark & \boldxmark \\
    \rowcolor{gray!10}
    MUSK & C & C & \boldxmark & \boldxmark & & Z/F/C & Z & \boldxmark & & Z & Z & \boldxmark & C & & C & C\\
    BEPH & Z/F/C & C & \boldxmark & \boldxmark & & C & \boldxmark & \boldxmark & & \boldxmark & \boldxmark & \boldxmark & \boldxmark & & \boldxmark & \boldxmark\\
    \rowcolor{gray!10}
    Hibou & C & \boldxmark & \boldxmark & \boldxmark & & C & \boldxmark & C & & \boldxmark & \boldxmark & \boldxmark & \boldxmark & & C & \boldxmark \\
    mSTAR & Z/F/C & C & \boldxmark & \boldxmark & & \boldxmark & \boldxmark & \boldxmark & & \boldxmark & \boldxmark & C & \boldxmark & & C & C\\
    \rowcolor{gray!10}
    GPFM & C & C & \boldxmark & \boldxmark & & C & Z & \boldxmark & & \boldxmark & \boldxmark & C & C & & C & \boldxmark \\
    Virchow2 & \boldxmark & \boldxmark & \boldxmark & \boldxmark & & \boldxmark & \boldxmark & \boldxmark & & C & \boldxmark & \boldxmark & \boldxmark & & \boldxmark & \boldxmark\\
    \rowcolor{gray!10}
    MADELEINE & F & C & \boldxmark & \boldxmark & & \boldxmark & \boldxmark & \boldxmark & & \boldxmark & \boldxmark & \boldxmark & \boldxmark & & \boldxmark & F/C\\
    Phikon-v2 & F/C & \boldxmark & \boldxmark & \boldxmark & & \boldxmark & \boldxmark & \boldxmark & & \boldxmark & \boldxmark & \boldxmark & \boldxmark & & F/C & F/C \\
    \rowcolor{gray!10}
    TITAN & Z/F/C & C & Z & \boldxmark & & C & \boldxmark & \boldxmark & & Z & Z & C & \boldxmark & & C & C\\
    KEEP & Z & \boldxmark & \boldxmark & Z & & Z & \boldxmark & \boldxmark & & Z & Z & \boldxmark & \boldxmark & & \boldxmark & \boldxmark \\
    \rowcolor{gray!10}
    THREADS & F/C & C & Z & \boldxmark & & \boldxmark & \boldxmark & \boldxmark & & \boldxmark & \boldxmark & \boldxmark & \boldxmark & & C & F/C\\
    \bottomrule
    \end{tabular}
    \caption{Comparison of the evaluation tasks between different PFMs. Abbreviations used: Zero-shot (Z), Few-shot (F), Complete (C).}
    \label{tab:benchmark}
\end{table*}
\section{Evaluation Tasks for the Foundation Model}\label{sec:benchmark}
Development and evaluation constitute the two fundamental pillars of PFMs. The evaluation tasks of PFMs can be systematically categorized into four aspects: 1) slide-level tasks; 2) patch-level tasks; 3) multimodal tasks; and 4) biological tasks. A comparative analysis of these evaluation tasks is presented in \cref{tab:benchmark}, providing practitioners with comprehensive criteria for model selection based on real-world applications.

\textbf{Slide-level Tasks} encompass analytical tasks that utilize WSIs as primary input or output modalities. These tasks include WSI classification (Cls.), survival prediction (Surv.), WSI retrieval (Retri.), and WSI segmentation (Seg.). While survival prediction methodologically represents a classification employing specialized loss functions, its clinical application differs from standard WSI classification: the latter primarily serves for diagnosis, while the former addresses prognosis. This category is the cornerstone of CPath, enabling automated diagnosis directly from WSIs with minimal manual intervention. Consequently, the majority of methods have prioritized experimental validation in this domain.

\textbf{Patch-level Tasks} comprise analytical tasks on patches as inputs or outputs, including patch classification (Cls.), patch-to-patch retrieval (P2P), and patch segmentation (Seg.). These tasks effectively evaluate the efficacy of the extractor, as they operate independently of additional aggregators. 

\textbf{Multimodal Tasks} are tasks that evaluate multimodal capabilities of PFMs. These tasks encompass cross-modal retrieval, \ie, image-to-text (I2T) and text-to-image (T2I) retrieval, report generation (RG), and visual question answering (VQA).
RG in our survey includes both RG and image captioning, distinguished by their input: RG utilizes WSIs to generate clinical documentation, while captioning produces concise descriptions from patches. The increasing emphasis on these tasks reflects the clinical reality that pathologists integrate multimodal data in the decision-making process.

\textbf{Biological Tasks} focus on biomarker detection, including genetic alteration (GA) and molecular prediction (MP). Genetic alteration includes both mutation prediction and genetic alteration, as both predict gene mutation status. Molecular prediction targets the prediction of molecular subtypes at the gene expression level, representing a distinct biomarker from genetic alteration. While these tasks can be fundamentally categorized as classification problems at either slide or patch level, their clinical applications and biological implications warrant their classification as a separate analytical category. One recently-proposed task is molecular prompting~\cite{vaidya2025thread}, which aims to perform clinical tasks with canonical molecular profiles without requiring any task-specific model development in a similar manner to text prompting.

While the extensive scope of our evaluation tasks precludes exhaustive evaluation by any single model, several methods, notably CONCH, UNI, MUSK, GPFM, and TITAN, provide excellent evaluation benchmarks across multiple training paradigms, including zero-shot, few-shot, and complete supervised learning, thereby providing more holistic insights into model capabilities and generalization potential. 

\section{Future Directions}\label{sec:challenges}
PFMs constitute an emerging paradigm with transformative potential. Future research directions bifurcate into two primary domains: effective PFM Development and Utilization.

\subsection{Foundation Model Development}
\textbf{Pathology-specific Methodology} design is essential for PFMs that effectively capture the unique characteristics of pathology data. Most PFMs are pretrained using algorithms originally developed for natural images, neglecting critical aspects of pathology images, as detailed in \cref{sec:taxonomy}; therefore, there is an urgent need for algorithms designed to accommodate these challenges. This deficiency extends to multimodal pretraining as well, where CLIP and CoCa are employed without customization, resulting in the omission of inherent features of pathology and related data, including genomics and reports, that are vital for comprehensive analysis.

\textbf{End-to-end Pretraining} is critical to achieve optimal performance for PFMs. Current PFMs adopt a two-stage pretraining paradigm: extractors are trained independently, followed by the aggregator with the extractor frozen. Evidence suggests this complicates optimization, highlighting the need for end-to-end pretraining of PFMs, which poses significant challenges in CPath, as transitioning away from MIL requires developing extremely sophisticated and efficient architectures and algorithms capable of simultaneously integrating local and global pathology information for gigapixel images.

\textbf{Data-Model Scalability} is a critical direction, as performance improvements continue to demonstrate logarithmic and sub-logarithmic scaling with model and data volume, respectively, without yet reaching a clear plateau. This domain presents four sub-directions: 1) examining the relative importance of WSI and patch quantity, particularly when considering diversity, a complex concept that is yet widely acknowledged as an indicator of high-quality data; 2) exploring efficient algorithms, due to the rapid expansion in both datasets and models, evident in transitions from CONCH (ViT-B) to CONCHv1.5 (ViT-L) and from UNI (ViT-L) to UNI2 (ViT-H); 3) addressing the data-model scale mismatch problem for the aggregator, detailed in \cref{sec:scope}; and 4) optimizing model scale, since the giant model size poses substantial deployment challenges in both hospital and academic settings.

\textbf{Federated Learning with Efficiency} is essential for addressing the challenges associated with collecting massive-scale datasets across multiple institutions while preserving patient privacy, as few institutions can feasibly collect WSIs at the million-scale alone. However, current research in this area remains limited; for instance, HistoFL \cite{lu2022federated} has demonstrated improved performance, yet this benefit comes at the cost of significantly increased computational overhead. As PFMs continue to grow in size, scaling federated learning further exacerbates these challenges. Consequently, there is an urgent need to develop more efficient, privacy-protected methods in such large-scale cross-institutional collaborations.

\textbf{Model Robustness} addresses critical challenges in multi-institutional data curation. The acquisition of data from various sites inevitably introduces technical heterogeneity across scanning equipment specifications, image magnification levels, and staining protocols, resulting in significant data variations that embed site information \cite{de2025current}. These disparities undermine training stability and model generalizability; recent work shows that most models encode site information more strongly than biological signals \cite{de2025current}. These issues will be further exacerbated in federated learning under non-IID data distributions. Consequently, developing more robust algorithms and robustness evaluation metrics for PFMs is a critical research imperative.

\textbf{RAG-enhanced Pathology VLM} is a trending paradigm worth investigating. Contemporary trends in \textbf{L}arge \textbf{L}anguage \textbf{M}odels (LLMs), such as Llama \cite{grattafiori2024llama}, with the prevalence of BERT-based architectures \cite{devlin2019bert} in current multimodal PFMs suggest the potential utility of integrating LLMs with ViT architectures. Furthermore, given the demonstrated efficacy of \textbf{R}etrieval-\textbf{A}ugmented \textbf{G}eneration (RAG) \cite{gao2023retrieval} in LLMs and the critical need for domain-specific expertise in pathology, RAG methodology offers promising directions for representation learning in pathology VLMs. This approach transcends the limitations of existing methods such as RudolfV, which relies primarily on clustering techniques for pathologist knowledge integration, providing a potentially more sophisticated framework for incorporating domain expertise.

\subsection{Foundation Model Utilization}
\textbf{Effective Adaptation} of PFMs to downstream tasks is a critical research direction in their utilization, as these models are predominantly trained on large-scale heterogeneous datasets, resulting in general-purpose features rather than task-specific ones required for optimal performance. To address this limitation, effective adaptation methodologies are essential for task-specific optimization. The significance of this domain alignment challenge parallels the established paradigm of adapting conventional architectures, such as ResNet-50, to specialized domains like pathological image analysis, albeit with varying degrees of complexity and scope.

\textbf{Model Maintenance} constitutes a critical research domain in the context of PFMs, given the substantial computational resources required for their initial training. The potential diminishment of model performance due to novel diseases, tissue heterogeneity, or technological advancements necessitates efficient maintenance strategies to preserve model utility. Continual learning \cite{wang2024comprehensive,yu2024recent} represents a promising approach for maintaining PFM effectiveness, as it circumvents the necessity for model retraining by learning on newly observed instances. This approach significantly reduces the required computational overhead while ensuring the model remains current with the evolving clinical, disease, and technological developments.

\section{Conclusion}\label{sec:conc}

This survey presents a systematic analysis of the current Pathology Foundation Models through our proposed hierarchical taxonomy and comprehensive evaluation framework. Although the PFMs demonstrate significant advances in computational pathology, critical technical challenges merit further investigation. We delineate key directions that are worth exploring and might be instrumental in advancing both the theoretical foundations and practical applications of PFMs.

\appendix

\section*{Acknowledgments}
The authors would like to sincerely thank Professor Irwin King from the Department of Computer Science and Engineering, the Chinese University of Hong Kong, for his active involvement in the conception and development of this paper. Due to the one-submission-per-author policy, his name could only appear in the acknowledgments. The work described in this paper was partially supported by the Research Grants Council of the Hong Kong Special Administrative Region, China (CUHK 2410072, RGC R1015-23).

\bibliographystyle{named}
\bibliography{main}

\end{document}